\documentclass[runningheads]{llncs}
\usepackage{graphicx}
\usepackage{multirow}
\usepackage{array}
\usepackage{ragged2e}
\begin{document}
\title{Domain Controlled Title Generation with Human Evaluation}
%
%
\author{Abdul Waheed\textsuperscript{\orcidID{0000-0002-7229-3207}} \and Muskan Goyal\textsuperscript{\thanks{Corresponding Author}\orcidID{0000-0003-3643-7311}} \and \\   Nimisha Mittal\textsuperscript{\inst{*}\orcidID{0000-0002-7178-2017}}\and Deepak Gupta\textsuperscript{\orcidID{0000-0002-3019-7161}}} 
%

\institute{Computer Science Department\\ Maharaja Agrasen Institute of Technology \\
\email{e.abdul@protonmail.com, goyalmuskan1508@gmail.com\inst{*}, nimishamittal1999@gmail.com\inst{*},deepakgupta@mait.ac.in}\\}
%
\maketitle              

\begin{abstract}
We study automatic title generation and present a method for generating domain-controlled titles for scientific articles. A good title allows you to get the attention that your research deserves. A title can be interpreted as a high-compression description of a document containing information on the implemented process. For domain-controlled titles, we used the pre-trained text-to-text transformer model and the additional token technique. Title tokens are sampled from a local distribution (which is a subset of global vocabulary) of the domain-specific vocabulary and not global vocabulary, thereby generating a catchy title and closely linking it to its corresponding abstract. Generated titles looked realistic, convincing, and very close to the ground truth. We have performed automated evaluation using ROUGE metric and human evaluation using five parameters to make a comparison between human and machine-generated titles. The titles produced were considered acceptable with higher metric ratings in contrast to the original titles. Thus we concluded that our research proposes a promising method for domain-controlled title generation.

\end{abstract}

\section{Introduction}

The analysis of literature is a crucial practice for researchers to determine relevant publications for the research topic. The title of the research papers becomes really significant due to the availability of scientific papers in abundance. Researchers can automatically determine the importance of a paper by its title rather than reading the whole document \cite{1,2,3}. The accuracy of the title influences the number of potential readers and therefore the number of citations \cite{1,4}. That’s why it’s important for the researchers to produce a good title, however, people spend very little time on it \cite{3}. This leads to non-informative headings that do not take the entire content of the scientific paper into account.\par

To establish a title for a scientific paper, one has to understand the objective and characteristics of the paper, so that the meaning of the paper can be distilled into the title with only a few words. A suitable title for a particular paper must convey the core meaning of the paper briefly. The automatic title generation task has traditionally been closely linked to the conventional summarization technique because it can be considered as scientific paper compression to represent its content \cite{5,6,7}. For a given paper, a short and concise description, conveying the complete details of the text in just a few words, must be made. The task cannot, therefore, be regarded as simple.\par
It is crucial to identify the type of information/domain of each text unit to generate a suitable title. The performance of automatic title generation can become unsatisfactory due to the differences between the domains of scientific papers. Such differences include diverse vocabulary, different forms of grammar, and various ways of expressing identical concepts. Therefore, it is important to use domain-specific vocabulary to generate the title for scientific papers. Although automated title creation is not capable of replacing an author's skills in creating a title, it is helpful to propose a title.\par
In this research, we propose to utilize the text-to-text transformer model (T5 model) for domain-controlled title generation. The text-to-text transformer model converts all the NLP tasks to a single text-text format where text is used as input and output. This formatting allows the T5 model to perform a number of tasks into its framework, including the summarization task. For domain control, we used an additional token method which is discussed in Section \ref{section:3.3}. Since an abstract of a scientific paper describes the author’s work and presents all key arguments and relevant findings \cite{8}, we used it to generate a suitable domain-controlled title for papers. Several researchers have worked on automatic title generation. We examine some of these works below. 

\subsection{Related Work}
Liqun and Wang proposed a method called DTATG in \cite{9} to produce titles. DTATG was an unsupervised method that produced syntactically correct titles very easily. This method used sentence segmentation to draw a limited number of central sentences that express the core meanings of the text. DTATG created a dependency tree for each of these sentences and extracted several branches for trimming purposes with a dependency tree compression model. The authors also designed a title test to figure out if the trimmed sentence can be used as a title. The title with the top-ranked score was selected as the final title. DTATG created titles similar to human-generated titles. However, DTATG was constrained to the use of central phrases.\par
In \cite{10}, Putra et al. proposed a method for the creation and classification of rhetorical corpus structures. The experiment was directly integrated into the role of automatic title generation for scientific papers. Each sentence was categorized into one of the three groups: OWN\_MTHD (method), AIM (purpose), and NR (not relevant). Firstly the abstract of the scientific papers was annotated with the corresponding category. Then features like rhetorical patterns and formulaic lexicon statistics were decided and analyzed for classification purposes. Finally, several supervised learning algorithms were utilized to build the classification model. The models were tested using 10-fold cross-validation and resulted in a weighted average F-measure between 0.70-0.79.\par
Liu, Wei, et al. \cite{11} proposed a sentence2vec-enhanced Quality-Diversity Automated Summarization (QDAS) model and attempted to implement transfer learning for the Wikipedia title Generation task. Summaries from paragraphs were derived by extractive summarization method and sequence labeling data was provided to the model for title generation. The system involved only general preprocessing, such as sentence splitting and word segmentation, and could be implemented in a multilingual environment. The authors fine-tuned the BERT based CRF model for Wikipedia article title generation.\par 
Chen et al. \cite{12} presented a method for domain adaptation using artificial titles for title generation. The authors discussed the strategies for modifying the encoder-decoder model for text generation from a domain marked source to an unlabeled domain target. Sequential training was utilized in order to capture the unlabeled target domain's grammatical form. For the title generation task, an encoder-decoder RNN model with domain control was used. The encoder collected data from the source and the decoder learned to construct summary captions. The source data and unlabeled target domain information were encoded as their definition representations by a bidirectional LSTM, and the domain classifier attempted to learn to distinguish between the representations of two domains.\par 
In \cite{13}, Gehrmann, Sebastian et al. proposed a method to generate titles for short sections of long documents. The authors aimed at designing techniques in a low-resource system for section title generation. They first picked the most influential sentence and then performed a removal-based compression on an extractive pipeline. The Semi-Markov Conditional Random Field compression method was utilized. The method depended upon unsupervised textual representations such as ELMo or BERT to eliminate the need for the design of the complex encoder-decoder.\par 
Most of the above works are centered on how relevant terms can be extracted in the article for the title output without recognizing the domain of the text. The domain offers a category of material that reflects the communication purpose conveyed by a paper to the reader. It is alleged that the sentence domain in the document could boost the effectiveness of automatic title generation by including a certain sort of information in the document title. In this research, we consider domains in the form of information types communicated by abstracts of scientific papers. Also, we used a transformer-based architecture model (T5 model) instead of using sequence-to-sequence models (RNN, LSTM)  to perform automatic domain-controlled title generation. RNN/CNN handle sequences word-by-word sequentially which is an obstacle to parallelize. Transformers remove recurrence and convolution entirely and substitute them with self-attention to assess the dependencies between inputs and outputs. Transformer achieve parallelization by replacing recurrence with attention and encoding the symbol position in sequence. This in turn leads to significantly shorter training time and better performance. 

\subsection{Contributions}
While considering the above works, we have proposed a domain-controlled automatic title generation method. The key contributions to this paper are mentioned as follows:\par 
\begin{enumerate}

\item Propose a method for automatic domain-controlled title generation using transformers.
\item Provide meta-information about the domain using the additional token technique.
\item Compare the titles generated by the model with human-generated titles by performing the human evaluation.
\end{enumerate}
The remaining exhibition is the following. The background is presented in Section \ref{section:2}. The proposed model is discussed in Section \ref{section:3}. Section \ref{section:4} deals with the simulation model, followed by Section \ref{section:5} with results and discussion. Section \ref{section:6} addresses the conclusions and possible future directions.

\section{Background}\label{section:2}
Natural Language Processing (NLP) is a facet of Artificial Intelligence and is responsible for computers and machines to understand, interpret, and further incorporate human languages. NLP is based on different disciplines, like computer science and machine-translated linguistics. It further attempts to narrow the communication gap between computers and humans.
\subsection{Attention}\label{section:2.1}
Attention \cite{14} proposed in machine translation alleviates the bottleneck of seq2seq learning. It allows us to look at hidden states of the source sequence and they are parsed as an additional input in the form of weighted average to the decoder. It is not restricted to input sequence and the concept of self-attention can be incorporated to analyze surrounding words to obtain contextually sensitive word representations. Transformer architecture [15] consists of various self-attention layers.

\subsection{Pretrained language models}\label{section:2.2}
The pre-trained word embeddings initialize only the first layer of the model and the embeddings are generally context-agnostic. Pretraining always learns the important parameters of deep neural networks and are further fine-tuned for various tasks. They were proposed in 2015 \cite{16} but their benefits across various tasks were recently observed. Pretraining involves improved model initialization that boosts generalization performance and convergence speed for the tasks. Being language models, they can solely rely on unlabeled corpora, hence they are beneficial when labeled data is scarce.

\subsection{Transformer architecture}\label{section:2.3}
The Transformer \cite{15} is a model architecture that solves the problem of parallelization by integrating self-attention techniques with convolutional neural networks. Before the Transformer was introduced, Convolutional Neural Networks (CNN) and Recurrent Neural Networks (RNN) were used for sequence transduction. Similar to RNNs and CNNs, there is an encoder and decoder in the transformer where each block includes a multi-head attention block, skip connection, and normalization layer followed by a feed-forward block. With the help of multi-head attention and positional embedding, the transformer network has the capability of processing the data parallely. Often, transformer-based models contain stacked encoders and decoders. \par 
The self-attention mechanism in this model is responsible for correlating each input token regardless of their position. Initially, all the words are represented using high dimensional vectors or embeddings. This process is performed in the bottom-most encoder that is subsequently passed to all the succeeding encoders which perform self-attention to generate new representation considering contextual information. The decoder having similar architecture operates likewise but generates one word at a time by attending the encoded contextualized representation and previously generated tokens.

\section{Proposed Method}\label{section:3}
This section presents the methodology and architecture used in the research. A detailed overview of the model and the process used to make the overall concept comprehensible has been given. The present study discusses title generation as a summarization process. Our suggested architecture comprises three modules integrating a primary pipeline for summarization: pre-processing, training of the T5 model, and generation of the title (summary). The flowchart of the proposed method is shown in Fig.~\ref{fig1} below. Each module is defined in detail in the following sections.
\begin{figure}
\includegraphics[width=\textwidth]{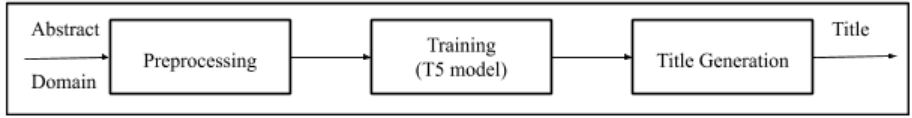}
\caption{Flowchart of title generation using T5. Preprocessing involves mainly tokenization which is based on subword tokenization.} \label{fig1}
\end{figure}

\subsection{Pre-processing: Subword Tokenization}\label{section:3.1}
The tokenization of subwords is a strategy that embraces uncertainty in data. Subword tokenization breaks the text into subwords. For instance, terms like smartest are segmented as smart-est, lower as low-er, etc. Transformer based models depend on subword tokenization algorithms for vocabulary generation. This method divides OOV(out of vocabulary) terms into subwords and depicts the term in these subwords. The length of input and output phrases is less than the tokenization of the character after subword tokenization. This method uses the most frequently occurring character or sequence iteratively.

\subsection{T5 Model}\label{section:3.2}
T5: Text-to-text-Transfer-Transfer model was proposed by Colin Raffel et al. \cite{17} to reframe all NLP-based activities to a single text-to-text format where text strings are both input and output. This formatting enables the T5 model to cast various tasks into this framework including machine translation, classification task, regression task, and summarization task. This framework uses the same model, hyperparameters, and loss function on the tasks mentioned.\par The model architecture is similar to standard vanilla encoder-decoder transformers with similar computational cost and considering half parameters. The input sequence is initially mapped to the embedding sequence and transferred to the encoder afterwards. The encoder comprises a stack of blocks, and each block consists of two subcomponents: a self-attention layer and a feed-forward network. Normalization of the layer is introduced to each subcomponent’s input. Normalization only includes rescaling of the activations and no additive bias is introduced. This is followed by a residual skip connection that adds the subcomponent's input to its output. This model considers a different position embedding scheme and these are passed as parameters across all layers of the model. \par 
The structure of the decoder is similar to the encoder except it incorporates the standard attention process after the self-attention layer encoder’s output. The self-attention in the decoder is a type of auto-regressive or causal self-attention that only enables the algorithm to attend to previous outputs. The final decoder block output is fed into a dense layer with a softmax output, the weights of which are shared with the input embedding matrix. Hence, the T5 model is similar to the original Transformer with an exception of the removal of Layer Norm bias and using different position embedding methods.

\subsection{Additional token technique for domain control}\label{section:3.3}
Domain control or often called adaptation is an important research area in the processing of natural languages. The arXiv dataset has a category/domain label corresponding to each paper. By controlling the domain, title tokens will be sampled from the local distribution of the domain-specific vocabulary (which will be a subset of global vocabulary) rather than global vocabulary. The produced title will therefore be attractive and realistic, closely associated with its corresponding abstract. \par 
Now, it would be cumbersome and time-consuming to train one separate model for each type of domain. The solution to this problem is to add a reference to the desired domain to the input text. The reference acts as a control tag specific to each type of domain. This helps the model to learn the relation between the control tags and the following text and only a preferred control tag must be defined by the user to specify the type of text that is to be generated. This control tag is the meta-information about the domain that we provide to the model. The tags are supplied using the additional token method technique.\par 
The additional token method \cite{18} requires the addition of an artificial token at the end of each abstract which enables the model to pay attention to the domain of each title and abstract. The model reads the abstract with the associated domain tag @DOMAIN-NAME. Domain tags are already defined in the ArXiv metadata file, corresponding to each paper. Although simple, this approach has proved to be successful \cite{19,20}. Note that the T5 model does not include any information about the domain. It includes domain information through this additional token approach. Now, we need to maximize the probability of title using abstract and domain.
\par 
\noindent Domain Controlled, 
\begin{equation}
    loss = -log\sum_{t=1}^{|y|} P(y_t^*|y_{<t}^*,x,d) 
\end{equation}
 
\noindent Without domain control,
\begin{equation}
    loss = -log\sum_{t=1}^{|y|} P(y_t^*|y_{<t}^*,x) 
\end{equation}

\noindent where,\par 
\noindent \(y_t^* = y^{<1>},y^{<2>},...,y^{<n>}\) denotes tokens in generated title \par
\noindent \(x = x^{<1>},x^{<2>},...,x^{<m>}\) denotes tokens of the abstract \par 
\noindent \(d\) denotes domains (examples astro-ph: Astrophysics, cs.CC: Computational Complexity, cs.CL: Computation and Language)

\subsection{Training Procedure}\label{section:3.4}
Aforementioned, a T5 model takes input and produces output in text format. A unique prefix tag is used to define and train the model on different tasks. The format of input text passed to the model consists of three parameters, prefix, input text, and target text. This makes training of the model simple as the model only requires a change in the prefix tag to run a specific task. The format of input can be stated as, \({<}\)prefix\({>}\):\({<}\)input\_text\({>}\). \par
\begin{figure}
\includegraphics[width=\textwidth]{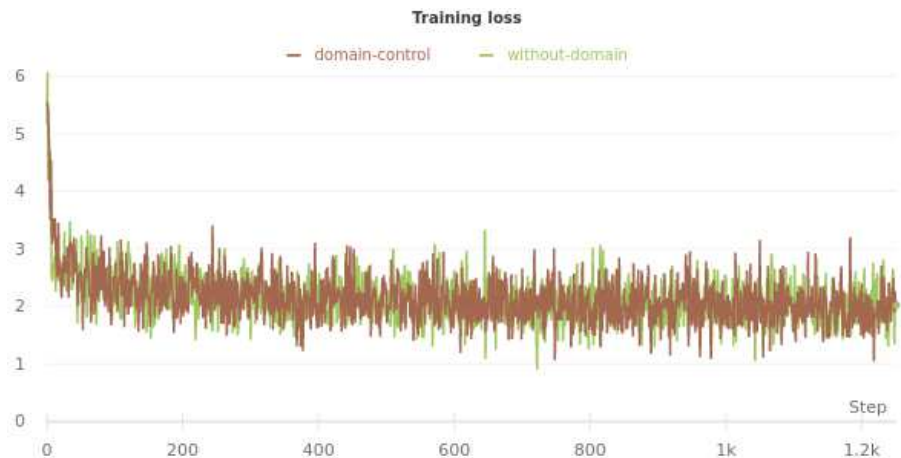}
\caption{Training Loss} \label{fig2}
\end{figure}

 
The first step is to feed the text input into a layer of word embedding that generates a vector representation of each word. Further, positional encoding is used to inject positional data into the input embeddings. The Encoder layer present transforms input sequences to abstract continuous representation. The decoder is responsible for producing text sequences that are capped off with a linear and a softmax layer. Moreover, the decoder consists of a start token, a list of previous outputs that play the role of inputs along with encoder outputs containing attention information. It stops decoding when a token is produced as an output.
\\\indent In our implementation, we used Beam Search with beam size 4 as the decoding method because it eliminates the likelihood to miss hidden word sequences with high probability.  For instance, consider the abstract and title pair illustrated below. This abstract is further passed to the T5 model after appending the appropriate domain tag as represented to predict domain controlled title of the scientific paper. The training loss graph of the model for domain-control and without-domain control is shown in Fig.~\ref{fig2}.
\par 

\justifying
\noindent
\textbf{Abstract:}
\\
\textit{This paper proves that labelled flows are expressive enough to contain all process algebras which are a standard model for concurrency. More precisely, we construct the space of execution paths and of higher-dimensional homotopies between them for every process name of every process algebra with any synchronization algebra using a notion of labelled flow. This interpretation of process algebra satisfies the paradigm of higher dimensional automata (HDA): one non-degenerate full $n$-dimensional cube (no more no less) in the underlying space of the time flow corresponding to the concurrent execution of $n$ actions. This result will enable us in future papers to develop a homotopical approach to process algebras. Indeed, several homological constructions related to the causal structure of time flow are possible only in the framework of flows.}
\\
\textbf{Actual Title:}\\
\textit{Towards a homotopy theory of process algebra}

\par
\indent The T5 model reads the Abstract appended with the appropriate domain tag (@DOMAIN-NAME) of the scientific paper.
\par

\justifying
\noindent
\textbf{Input(Abstract):}\\
\textit{This paper proves that labelled flows are expressive enough to contain all process algebras which are a standard model for concurrency. More precisely, we construct the space of execution paths and of higher-dimensional homotopies between them for every process name of every process algebra with any synchronization algebra using a notion of labelled flow. This interpretation of process algebra satisfies the paradigm of higher dimensional automata (HDA): one non-degenerate full $n$-dimensional cube (no more no less) in the underlying space of the time flow corresponding to the concurrent execution of $n$ actions. This result will enable us in future papers to develop a homotopical approach to process algebras. Indeed, several homological constructions related to the causal structure of time flow are possible only in the framework of flows. \textbf{@domain: math.AT math.CT}}
\\
\textbf{Output(Predicted Title):}\\
\textit{Labelled flows and homotopical approach to synchronization algebras}

\par

\par
\section{Simulation Model}\label{section:4}
\subsection{ Simulation Setup }\label{section:4.1}
The entire model is trained and tested on an Nvidia Tesla P100 GPU with 16GB GPU VRAM using python language. The overall parameters of the model and dataset are defined below.
\subsection{Parameters}\label{section:4.2}
The T5 model used various parameters in the training and testing process. By adjusting these parameters, the accuracy of the model can be determined. The maximum sequence length of the input (abstract) is set to 512 and the maximum sequence length of output (title) is set at 20.  For both training and testing, a batch size of 8 is used . The model is trained with a learning rate of 1e-5 for a total of 5 epochs. An evaluation occurred once for every 1000 training steps.

\subsection{Dataset}\label{section:4.3}
The model is trained on the popular ArXiv collection \cite{21}. ArXiv has served the general and research community with free access to scholarly publications from the major fields of physics to several computer sciences disciplines and all that goes between them, including electrical engineering, computational biology, mathematics, and statistics. This large corpus of data provides substantial, but often daunting, depth. This dataset contains the original data from ArXiv. Since the complete dataset is very huge (with 1.7 million articles), it provides a JSON-format metadata file. For each paper, this file contains an entry with key features such as names of articles, authors, domains/categories, abstracts, complete text PDFs, etc. For this research, we used 1,10,000 samples from the dataset that is further divided into 90\% for training and 10\% for the purpose of validation.

\section{Results and Discussion}\label{section:5}
\subsection{Automated Evaluation}\label{section:5.1}
Automatic evaluation is performed for evaluating systems when the output is the text that is generally referred to as a sequence to sequence or string transduction problem. For evaluating the results, generated text from the model is compared with target (source) or reference text. There are various evaluation metrics, but this paper focuses on the ROUGE metric.\par
Recall-Oriented Understudy for Gisting Evaluation (ROUGE) \cite{22} is an intrinsic metric that is recall focused. This metric measures the number of n-grams or words present in the target text summaries appeared in the generated text by machines. There are three prominent types of Rouge scores including Rouge-N, Rouge-L, and Rogue-S.\par
The N-gram overlap between the machine summary and the target summary is represented in Rouge-N. Rouge-N contains Rouge-1 and Rouge-2, which in generated and goal text correspond to the contrast of unigrams and bigrams, respectively. The longest matching sequence of terms using LCS is measured by Rouge-L. Rouge-S focuses on skip-bigrams based co-occurrence statistics. This paper uses only Rouge-1 and Rouge-L metrics for comparison. The scores of the ROUGE metric can be seen in Table \ref{tab1}.

\begin{table}
\caption{Results of the automatic evaluation. ROUGE score with abstract + domain is significantly better than without a domain.
}
\centering
\label{tab1}
\begin{tabular}
{>{\centering\arraybackslash} p{0.2\textwidth} >{\centering\arraybackslash} p{0.2\textwidth} >{\centering\arraybackslash} p{0.2\textwidth} >{\centering\arraybackslash} p{0.2\textwidth}}
\hline
Model &  Data & ROUGE-1 & ROUGE-L\\[0.5ex]
\hline
\multirow{2}{1em}{T5} &  Abstract & 28.5 & 16.6\\[0.5ex]
& {\bfseries Abstract+Domain} & {\bfseries 31.5} & {\bfseries 21.6}\\[0.5ex]
\hline
\end{tabular}
\end{table}

We trained the T5 model for title generation in two different ways. In the first one, we generated the title directly from abstract text where we observed that the tokens in the generated title were not specific to the domain of the paper. Hence, to address this issue we controlled the domain using additional token technique and with this approach we noticed significant improvement. It can be inferred from Table \ref{tab1} that ROUGE-1 and ROUGE-L scores obtained for domain controlled abstracts (Rouge-1: 31.5 and Rouge-L: 21.6) are considerably higher to the domain-less abstracts.

\subsection{Human Evaluation}\label{section:5.2}
The interest in analyzing NLG (Natural Language Generation) texts in recent years has grown by contrasting them to a corpus of human texts \cite{23}. As with other NLP fields, the benefits of automated corpus-based assessment are that it is theoretically both cheaper and faster than human-based assessment and is also repeatable. Corpus tests have been first used in the NLG, where the texts scanned from a corpus have fed the output of the parser into the NLG system and have then been compared to the original text of the corpus. In the NLG group, such corpus-based evaluations were often criticized. Reasons for criticism include the fact that the regeneration of a parsed text is not a true NLG task; that texts can vary considerably from corpus text but are still successful in achieving the communication purpose of the method and corpus texts often are not of good quality enough to shape a practical evaluation.\par
Human Evaluation is another approach to evaluate NLG (Natural Language Generation) based systems. This involves performing quality surveys of the generated output using human annotators. In this approach, generated results are presented to the people who assess the quality of the text on different criteria. Using intrinsic and extrinsic approaches \cite{23}, human evaluation of natural language generation systems can be performed. Intrinsic methods seek to test the performance properties of the system by asking participants about the fluency of the output of the system in a questionnaire. Extrinsic methods aim to measure the effect of the system by evaluating the degree to which the system accomplishes the overarching task for which it was created. Extrinsic testing is the most time-consuming and cost-intensive of all possible tests; hence it is very rare.\par
We conducted an intrinsic human evaluation with 40 participants. The academic background and field of study of each participant were also recorded. We prepared a form that paired 2 abstracts with the original and predicted title. The nature of the title (original or predicted) was not known to any of the participants. The participants judged and rated each title from 1(very bad) to 7 (very good). The judgments were based on various factors like coherence, relevance, fluency, semantic adequacy, and overall quality of the title. The aggregated mean score was first calculated for each factor, and then for each of the original and predicted titles. The values are shown in Table \ref{tab2}.\par
In the process of human evaluation, each factor has its own significance. Coherence determines whether or not the title is semantically meaningful, and relevance indicates how applicable the title is to the abstract. Two titles may equally effectively express the underlying intention while varying in fluency. Therefore, the titles were judged for their readability with fluency (able to understand the meaning of the title)  and semantic adequacy (the title is a sufficient representative of abstract) as its factors. It is often difficult for human annotators to differentiate between the different quality measures. This is why the overall quality of a title is evaluated directly. \par

The most suitable number of answers can rely on the task itself, but for most tasks, 7-point ratings are the best. Several studies suggest that 7-point ratings optimise reliability and discriminatory power \cite{24,25,26}. One 7-point likert question has 7 points for discrimination but 4 7-point likert questions have 4*7=28 discrimination points. Therefore, one 7-point likert rating was used for each of the 4 titles in human evaluation form.\par

\begin{table}
\caption{Human evaluation scores. Compared with human-scaled, T5 with domain generated titles are scored better than human-generated titles by human evaluators on almost all parameters.}\label{tab2}
\centering
\begin{tabular}
{>{\centering\arraybackslash} p{0.11\textwidth} >{\centering\arraybackslash} p{0.15\textwidth} >{\centering\arraybackslash} p{0.15\textwidth} >{\centering\arraybackslash} p{0.1\textwidth} >{\centering\arraybackslash} p{0.15\textwidth} >{\centering\arraybackslash} p{0.15\textwidth} >{\centering\arraybackslash} p{0.1\textwidth}}

\hline
Model &  Coherence & Relevance & Fluency & Semantic Adequacy & Overall Quality & Mean Score\\[0.5ex]
\hline
Human &  4.2 & 4.29 & 4.11 & 4.05 & 4.16 & 4.16\\[0.5ex]
{\bfseries T5 +  Domain} &  {\bfseries 4.53} & {\bfseries 4.63} & {\bfseries4.58} & {\bfseries 4.43} & {\bfseries 4.6} & {\bfseries 4.55}\\[0.5ex]
Scaled &  1.08 & 1.08 & 1.11 & 1.09 & 1.11 & 1.09\\[0.5ex]
\hline
\end{tabular}
\end{table}
\par

It is clearly evident from Table \ref{tab2} that the predicted title has higher values of evaluation metrics (coherence, relevance, fluency, semantic adequacy, and overall quality) in comparison to the original title. Overall, on normalizing the difference between the values of human-generated and machine-generated titles, the scaled value is in the range from 1.08 to 1.11. Moreover, the mean score of the title predicted using the proposed model is greater than human-generated titles. This distinctly depicts that the title produced using the T5 model and additional token is more appropriate and useful for the authors of scientific papers.

\section{Conclusion and Future work}\label{section:6}
In this research, we proposed the use of the T5 model for automatic domain-controlled title generation for scientific papers. We have used the ArXiv dataset for our research and analysis. Scholars, professional authors, students, and teachers may use this method. Sparseness is the biggest struggle of the title generation. For a document with several optional words, it is important to create a brief and succinct title. An abstract of a scientific paper contains the goal and method of the research, along with all the key findings. Therefore, we utilized a scientific paper’s abstract to produce the title for the paper. \par
We also identify the communication purpose of the authors bypassing the domain of the paper as the meta-information. The quality of the automatic title generated improves because the title tokens are sampled from the local vocabulary of the domain, rather than the global vocabulary. The titles generated from the model looked satisfactory and realistic.\par 
To analyze the results we used automated and human evaluation. For automated evaluation, we calculated ROUGE scores to prove that our approach leads to high performance in the generation of automatic domain-controlled titles. In the human-subject analysis, we used 40 participants with diverse reading capabilities and academic backgrounds. We noticed that our generated titles have higher values of evaluation metrics (coherence, relevance, fluency, semantic adequacy, and overall quality) in comparison to the original title.\par
Although the abstract is a brief representation of the paper, writers may have written the abstract hastily. Therefore, it is suggested to use the complete paper text along with more refined weighing, title generation, and selection methods, in order to better capture the prominent tokens of a document in a particular domain. Another task that can be addressed in the future is to create new techniques that deliver titles which are more close to human generated titles.

\end{document}